\documentclass[10pt,a4paper,twocolumn]{article}
\usepackage[T1]{fontenc}
\usepackage{times}
\usepackage[textwidth=170mm,textheight=235mm,columnsep=10mm]{geometry}
\usepackage{amsmath,amssymb}
\usepackage{tikz}
\usepackage{latexml}
\iflatexml
\let\figurefont\relax
\let\figurefontsmall\relax
\else
\font\figurefont=DejaVuSerif-tlf-t1 at 9bp
\font\figurefontsmall=DejaVuSerif-tlf-t1 at 8bp
\fi
\usepackage[hidelinks]{hyperref}
\title{Flash RNN-T: Memory-Efficient Joint Network and Transducer Loss Computation}
\author{Mahmoud Ashraf\\Tarteel AI\\\texttt{mahmoud.ashraf@tarteel.ai}}
\date{}
\begin{document}
\maketitle
\begin{abstract}
The recurrent neural network transducer (RNN-T) remains a practical choice for
streaming speech recognition, but training requires large intermediate tensors.
We present Flash RNN-T, a joint and loss implementation combining specialized
kernels with optional tiled recomputation over the full alignment space and
vocabulary.
Tiled and untiled Flash form the entire measured throughput-memory frontier
on medium and long workloads. On H100, untiled Flash achieves 24-39\% higher
throughput with 26-56\% less memory than the fastest baseline on these workloads
wherever it fits. Tiling reduces memory further through recomputation.
On medium, tiled Flash uses ninefold less memory than that baseline at 16\%
lower throughput when both use one backward call per batch.
On long, it fits where all tested alternatives with that backward schedule
run out of memory with 80 GB.
These comparisons use each implementation's lowest-memory configuration
reaching at least 95\% of its peak throughput.
\end{abstract}

\section{Introduction}\label{sec:1}

RNN-T remains a practical choice for streaming speech recognition, with recent
systems demonstrating accurate recognition under real-time constraints
\cite{xu2026chat,sirichotedumrong2026typhoon}.
Training an RNN-T requires marginalizing over alignments between an input sequence and a target
transcript \cite{graves2012transduction}. Its joint network
combines encoder and prediction network states at every valid position in the
alignment lattice, producing a distribution over the vocabulary at each position.
For a nonlinear joint, the resulting hidden states and vocabulary logits can
occupy substantially more memory than the scalar quantities needed by the
forward--backward recurrence \cite{li2019rnnt,braun2023samplewise}.
This makes the joint network and loss computation natural targets
for memory optimization.

Existing approaches reduce this memory cost by packing valid alignment positions
\cite{li2019rnnt}
and processing utterances individually \cite{braun2023samplewise}.
However, avoiding padded positions and processing utterances separately do not,
by themselves, bound the large intermediate tensors produced within a single
utterance. This motivates finer-grained control over their storage and lifetime.

We introduce Flash RNN-T, a joint and loss implementation combining specialized
GPU kernels with optional tiled checkpointing. Its tiled variant divides joint
network computation into groups of alignment positions, including within a
single utterance. For each group,
it computes the blank and target transition log probabilities needed by the
forward--backward recurrence, while discarding the much larger hidden states and
vocabulary logits.
To compute gradients without retaining these intermediates, we use activation
checkpointing \cite{chen2016checkpointing} to recompute
them one group at a time during backpropagation, trading additional computation
for control over their memory footprint. All valid alignment positions
and the full vocabulary are still evaluated, preserving the RNN-T objective.

We evaluate tiled and untiled Flash against existing implementations and examine
how specialization and recomputation contribute to their throughput--memory
tradeoffs. On the evaluated medium and long workloads, the two variants together
form the entire measured frontier on both GPUs, offering higher-throughput
untiled execution and lower-memory tiled execution.
Code, benchmark scripts and results are available online.\footnote{\url{https://github.com/MahmoudAshraf97/flash-rnnt-paper}}

\section{Background and related work}\label{sec:2}

\subsection{Joint network and alignment lattice}\label{sec:2.1}

For an utterance with $T$ encoder positions and $U$ target labels, the joint
network is evaluated at $T(U+1)$ alignment positions, including the empty
target prefix. Let $e_t$ and $p_u$ denote the encoder and prediction network
states projected to the joint hidden dimension, where $u$ counts labels
already emitted. The nonlinear joint \cite{graves2013deep} computes

\begin{equation}
h_{t,u}=\operatorname{Dropout}\!\left(\phi(e_t+p_u)\right),
\qquad
z_{t,u}=Wh_{t,u}+b,
\end{equation}

where $\phi$ is the activation function, $h_{t,u}$ is the hidden state, and
$W,b$ project it to logits $z_{t,u}$ over the $V$ vocabulary symbols,
including blank.

Although normalization requires all $V$ logits at each position, the RNN-T
forward--backward recurrence needs only the blank and valid next-target-label
log probabilities: $T(U+1)$ blank log probabilities and $TU$ target log
probabilities per utterance.
Conventional batched training retains the much larger joint intermediates for
gradient computation, so their memory cost accumulates across the batch
\cite{braun2023samplewise}.

\subsection{Memory-efficient transducer computation}\label{sec:2.2}

Li et al. \cite{li2019rnnt} reduce memory consumption by packing valid alignment positions
rather than padding every utterance to the batch's maximum lengths, removing
padded entries from joint hidden states and vocabulary logits. They compute
logit gradients directly, avoiding intermediate probability gradients, and
overwrite the probability tensor with those logit gradients to further reduce
storage.

GPU implementations also reduce intermediate storage and memory traffic.
Warp-transducer \cite{huangWarpTransducer} retains the raw logits and one normalization value per
alignment position, computing blank and target log probabilities directly
within the forward--backward recurrence instead of storing a full probability
tensor.
Apex \cite{nvidiaApex} supports packing and merged softmax/loss differentiation, while also
fusing joint addition, ReLU and dropout to avoid separate tensors for
intermediate sums and activations and reduce memory traffic.

Kuang et al. \cite{kuang2022pruned} instead reduce the number of alignment positions evaluated by the
expensive nonlinear joint. Their lightweight linear joint computes compact blank
and target log probabilities. These guide the selection of a narrow target range
at each encoder step, restricting the nonlinear joint's computation and
intermediate storage to that range.
This reduces memory and computation by restricting the alignment sum rather
than changing how the full sum is evaluated.

Without restricting alignments, Braun et al. \cite{braun2023samplewise} reduce the number of utterances
whose joint intermediates coexist in memory. They compute joint and loss
gradients utterance by utterance, accumulate gradients, and release utterance-local
tensors before propagating through the batched encoders. Consequently, the
large joint intermediates scale with an individual utterance rather than the
entire batch.

Flash RNN-T builds on packing, compact transition log probabilities and
checkpointing to control joint intermediate memory within individual utterances. We contribute
a GPU implementation that couples this finer-grained memory control with
specialized kernels, and quantify its throughput--memory tradeoffs against
existing implementations.

\section{Implementation}\label{sec:3}

\subsection{Tiled joint computation and checkpointing}\label{sec:3.1}

Tiled Flash RNN-T evaluates the joint network in small groups of alignment positions
called tiles. For each tile, it retains the compact blank and target transition
log probabilities while discarding the larger hidden states and logits, which
are recomputed during backpropagation.

\begin{figure}[tbp]
\centering
\iflatexml
\tikzset{every picture/.append style={scale=1.3,transform shape}}
\fi
\definecolor{ink}{HTML}{252525}
\definecolor{utteranceone}{HTML}{DBE9F3}
\definecolor{utterancetwo}{HTML}{F2E5CC}

\begin{tikzpicture}[
  x=1mm, y=1mm, draw=ink, text=ink,
  font=\small\figurefont,
  every node/.style={inner sep=0pt, outer sep=0pt, anchor=base,
                    yshift=-2.484375bp, execute at begin node=\vphantom{g}},
  bracket/.style={line width=.65bp, line cap=rect, line join=round}]
  \path[use as bounding box] (0,0) rectangle (80,22);
  \begin{scope}[overlay]
    \pgfmathsetmacro{\pitch}{76/17}
    \pgfmathsetmacro{\boundary}{2+8*\pitch}

    \foreach \i in {0,...,16} {
      \ifnum\i<8
        \fill[utteranceone] ({2+\i*\pitch},8) rectangle ++(\pitch,6);
      \else
        \fill[utterancetwo] ({2+\i*\pitch},8) rectangle ++(\pitch,6);
      \fi
    }
    \begin{scope}
      \clip (\boundary,8) rectangle (78,14);
      \foreach \i in {0,...,40} {
        \draw[line width=.35bp] ([xshift={8*\i bp}]-22,0) -- ++(80,80);
      }
    \end{scope}
    \foreach \i in {0,...,16}
      \draw[line width=.7bp] ({2+\i*\pitch},8) rectangle ++(\pitch,6);

    \foreach \a/\b/\name in {2/\boundary/1,\boundary/78/2} {
      \draw[bracket] ({\a+.25},15) -- ({\a+.25},16)
        -- ({\b-.25},16) -- ({\b-.25},15);
      \node at ({(\a+\b)/2},19.5) {Utterance \name};
    }
    \foreach \i in {6,12}
      \draw[line width=1.25bp, line cap=rect]
        ({2+\i*\pitch},7.5) -- ({2+\i*\pitch},14.5);
    \foreach \a/\b/\name in {0/6/1,6/12/2,12/17/3} {
      \draw[bracket] ({2+\a*\pitch+.25},7) -- ({2+\a*\pitch+.25},6)
        -- ({2+\b*\pitch-.25},6) -- ({2+\b*\pitch-.25},7);
      \node at ({2+(\a+\b)*\pitch/2},2.5) {Tile \name};
    }
  \end{scope}
\end{tikzpicture}
\caption{Packing two utterances into tiles with an illustrative limit of $R=6$
alignment positions.}
\label{fig:1}
\end{figure}
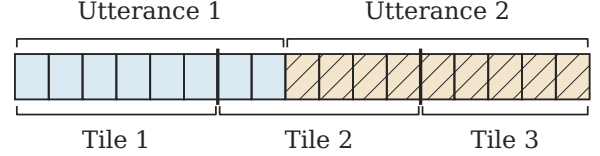

For a batch of $B$ utterances, let $T_b$ and $U_b$ denote the encoder and target
lengths of utterance $b$. We pack the $N=\sum_{b=1}^{B}T_b(U_b+1)$ valid alignment
positions into approximately balanced tiles of at most $R$ positions, which may
span utterance boundaries (\hyperref[fig:1]{Figure~\ref*{fig:1}}).

\begin{figure}[tbp]
\centering
\iflatexml
\tikzset{every picture/.append style={scale=1.3,transform shape}}
\fi
\definecolor{ink}{HTML}{252525}

\newcommand{\flowarrow}[4]{%
  \draw[line width=.75bp, line cap=round, line join=round]
    (#1) -- (#2);
  \draw[line width=.75bp, line cap=round, line join=round]
    ([xshift=#3bp,yshift=#4bp]#2) -- (#2)
    -- ([xshift=-#3bp,yshift=#4bp]#2);
}

\begin{tikzpicture}[
  x=1mm, y=1mm, draw=ink, text=ink,
  font=\small\figurefont,
  every node/.style={inner sep=0pt, outer sep=0pt, execute at begin node=\vphantom{g}},
  label/.style={anchor=base, yshift=-2.484375bp},
  small label/.style={anchor=base west, font=\footnotesize\figurefontsmall}]
  \path[use as bounding box] (0,0) rectangle (80,55);
  \begin{scope}[overlay]
    \draw[line width=.7bp, dash pattern=on 2.59bp off 1.12bp, line join=round]
      (20,44) -- (1,44) -- (1,24) -- (20,24);
    \draw[line width=.7bp, dash pattern=on 2.59bp off 1.12bp, line join=round]
      (32,44) -- (51,44) -- (51,24) -- (32,24);
    \foreach \y in {24,44}
      \draw[line width=.7bp, dash pattern=on 2.59bp off 1.12bp] (24,\y) -- (28,\y);

    \foreach \bottom/\height in {48/6,36/6,26/6,14/8,5/5}
      \draw[line width=.7bp] (4,\bottom) rectangle ++(44,\height);
    \node[label,yshift=.0625bp] at (26,51) {Projected states};
    \node[small label,yshift=2.58359375bp] at (55,51) {Checkpoint};
    \node[small label,yshift=-6.98984375bp] at (55,51) {inputs};
    \node[label] at (26,39) {Hidden states};
    \node[label,yshift=.125bp] at (26,29) {Logits};
    \node at (26,19.75) {Transition};
    \node at (26,16.25) {log probabilities};
    \node[small label,yshift=2.58359375bp] at (55,18) {Checkpoint};
    \node[small label,yshift=-6.98984375bp] at (55,18) {outputs};
    \node[label] at (26,7.5) {Loss};

    \foreach \upper/\lower in {48/42,36/32,26/22,14/10} {
      \coordinate (forwardstart) at ([yshift=-1bp]22,\upper);
      \coordinate (forwardend) at ([yshift=1.838525bp]22,\lower);
      \coordinate (gradientstart) at ([yshift=1bp]30,\lower);
      \coordinate (gradientend) at ([yshift=-1.838525bp]30,\upper);
      \flowarrow{forwardstart}{forwardend}{1.8}{3.6}
      \flowarrow{gradientstart}{gradientend}{1.8}{-3.6}
    }
    \coordinate (forwardstart) at (14,3.2);
    \coordinate (forwardend) at ([yshift=.838525bp]14,.8);
    \flowarrow{forwardstart}{forwardend}{1.4}{2.8}
    \coordinate (gradientstart) at (40,.8);
    \coordinate (gradientend) at ([yshift=-.838525bp]40,3.2);
    \flowarrow{gradientstart}{gradientend}{1.4}{-2.8}
    \node[small label,yshift=-2.203125bp] at (17,2) {Forward};
    \node[small label,yshift=-2.203125bp] at (43,2) {Gradients};
  \end{scope}
\end{tikzpicture}
\caption{Dashed enclosure marks intermediates discarded after each tile's
forward computation and recomputed during backward. Outputs from all tiles feed
the loss.}
\label{fig:2}
\end{figure}
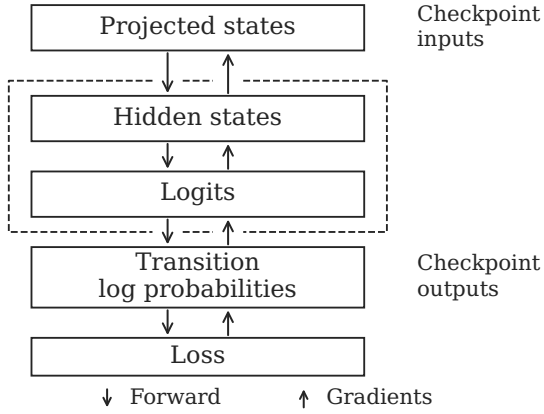

Once all tiles have produced their transition log probabilities, the
forward--backward recurrence computes the loss and its gradients with respect to
them. Because these gradients depend on all valid alignments of each utterance,
joint backpropagation must wait until this computation is complete. Each tile is
then recomputed and differentiated in turn, releasing its hidden states and
logits before processing the next tile (\hyperref[fig:2]{Figure~\ref*{fig:2}}).

The projected encoder and prediction network states remain outside the
checkpoint and are reused during recomputation. A saved seed per tile reproduces
the original dropout mask without retaining it. Tile recomputation runs inside
autograd backward, without changing the training loop.

\subsection{Specialized joint and loss kernels}\label{sec:3.2}

Within each tile, the joint kernel gathers the projected states for valid
alignment positions and fuses addition, activation and dropout, avoiding
intermediate tensors between these operations. Vocabulary projection uses
PyTorch's matrix multiplication separately. During backward, the joint kernel
fuses the dropout and activation derivatives with reductions along the encoder
and target axes, producing encoder and prediction network gradients from one
read of the hidden-state gradients.

The normalization kernel reads the full vocabulary logits but returns only the
blank and target log probabilities. During backward, it computes logit gradients
directly from the gradients of these two log probabilities, overwriting the
tile's logits rather than allocating a separate gradient tensor.

Copies between packed tiles and lattice tensors calculate their coordinates to
avoid storing an index for every position. The kernel identifies the utterance
from the packed boundaries and subtracts its starting offset, then obtains the
encoder and target coordinates through integer division and remainder by
$U_b+1$. This mapping places log probabilities into the lattice during forward
and gathers their gradients into tile order during backward.

\subsection{Storage and recomputation}\label{sec:3.3}

Let $H$ be the joint hidden dimension and $V$ the vocabulary size. Retaining
hidden states and logits for all $N$ packed positions requires
$O(N(H+V))$ storage. Checkpointing limits these tile-local tensors to
$O(\min(N,R)(H+V))$. The compact log probabilities and forward--backward buffers
remain rectangular, requiring $O(BT_{\max}(U_{\max}+1))$ storage, where
$T_{\max},U_{\max}$ are the maximum encoder and target lengths in the batch.
Projected states, parameters, gradients and metadata add further storage, so
$R$ controls the large joint intermediates rather than total device memory.

This reduction requires additional joint computation during backward, including
the vocabulary projection. Smaller tiles reduce peak workspace but increase the
number of kernel launches and partial gradient accumulations, creating a
throughput--memory tradeoff.

\section{Evaluation methodology}\label{sec:4}

We benchmark joint and loss forward and backward from projected encoder and
prediction network states, including vocabulary projection and gradients with
respect to joint parameters and projected states.

\subsection{Workloads}\label{sec:4.1}

We select short and medium workloads from the English subset of Granary
\cite{raokoluguri2025granary}
and a long workload from the linked long-form training split of TED-LIUM 3
\cite{fox2024longform}
to evaluate performance across realistic sequence lengths (Table~\ref{tab:1}). We use 2D bucketing
by audio duration and transcript length, with 30 duration buckets and two
transcript-length sub-buckets each
\cite{zelasko2025efficiency}.
Within each duration band, we select the batch closest to the band's median
duration, alignment-position count and padding fraction.

All experiments use $B=128$, $H=640$, a SentencePiece BPE \cite{kudo2018sentencepiece} vocabulary of $V=1024$
including blank, 80 ms encoder steps, and joint dropout of 0.1. We evaluate
primarily on NVIDIA H100 GPUs with 80 GB memory and repeat the experiments on
RTX 5090 GPUs with 32 GB to assess hardware dependence.

\begin{table}[tbp]
\centering
\caption{Representative workload batches.}
\label{tab:1}
\small
\setlength{\tabcolsep}{3pt}
\begin{tabular}{lrrrr}
\hline
\shortstack{Workload} & \shortstack{Mean\\duration\\(s)} & \shortstack{Encoder\\length $T$} & \shortstack{Transcript\\length $U$} & \shortstack{Valid\\alignment\\positions} \\
\hline
Short & 4.92 & 47--69 & 29--50 & 282,687 \\
Medium & 18.72 & 227--244 & 40--96 & 2,300,131 \\
Long & 69.62 & 851--885 & 325--458 & 41,122,726 \\
\hline
\end{tabular}
\end{table}

\subsection{Comparators and execution controls}\label{sec:4.2}

We compare tiled and untiled Flash RNN-T with Apex \cite{nvidiaApex},
NeMo Numba \cite{nvidiaNemoNumba} and Optimized
Transducer \cite{kuangOptimizedTransducer}, an implementation
of Li et al.'s loss optimizations \cite{li2019rnnt}.
The untiled variant omits tile checkpointing.

Joint inputs and computation use FP16; parameters, parameter gradients,
normalization and the forward--backward recurrence use FP32. The normalization
exception is Numba's FP16 shared-memory reduction intermediates; its denominators
remain FP32. Optimized Transducer additionally converts logits to FP32 to reuse
their buffer for FP32 gradients, with conversion costs included in timing and
memory.

Tiled Flash's tile limit doubles from 25k to 800k alignment positions. Other
implementations process the full batch or sub-batches of 1--64 utterances, doubling
the sub-batch size at each step. Retain keeps sub-batch graphs until a combined
backward call. Release applies Braun et al.'s sample-wise computation
\cite{braun2023samplewise} to sub-batches, differentiating
each immediately to accumulate parameter and projected-state gradients while
freeing intermediates.
Both compute the full-batch mean loss. Tiled Flash uses tile checkpointing in both
panels. Apex uses joint tile size 4 and joint/loss optimization setting 1.

\subsection{Measurement and numerical scope}\label{sec:4.3}

We report throughput as batch size divided by median latency and memory as
PyTorch's peak allocated memory, including inputs and parameters. Measurements
exclude compilation, autotuning and warm-up. Summary comparisons select each
implementation's lowest-memory configuration achieving at least 95\% of its peak
measured throughput, preserving near-peak speed while reducing memory. We validate
losses and gradients against reference calculations; the benchmark repository
documents the measurement protocol and correctness tests.

\section{Results}\label{sec:5}

\subsection{Comparison with existing implementations}\label{sec:5.1}

\begin{figure*}[!t]
\centering
\input{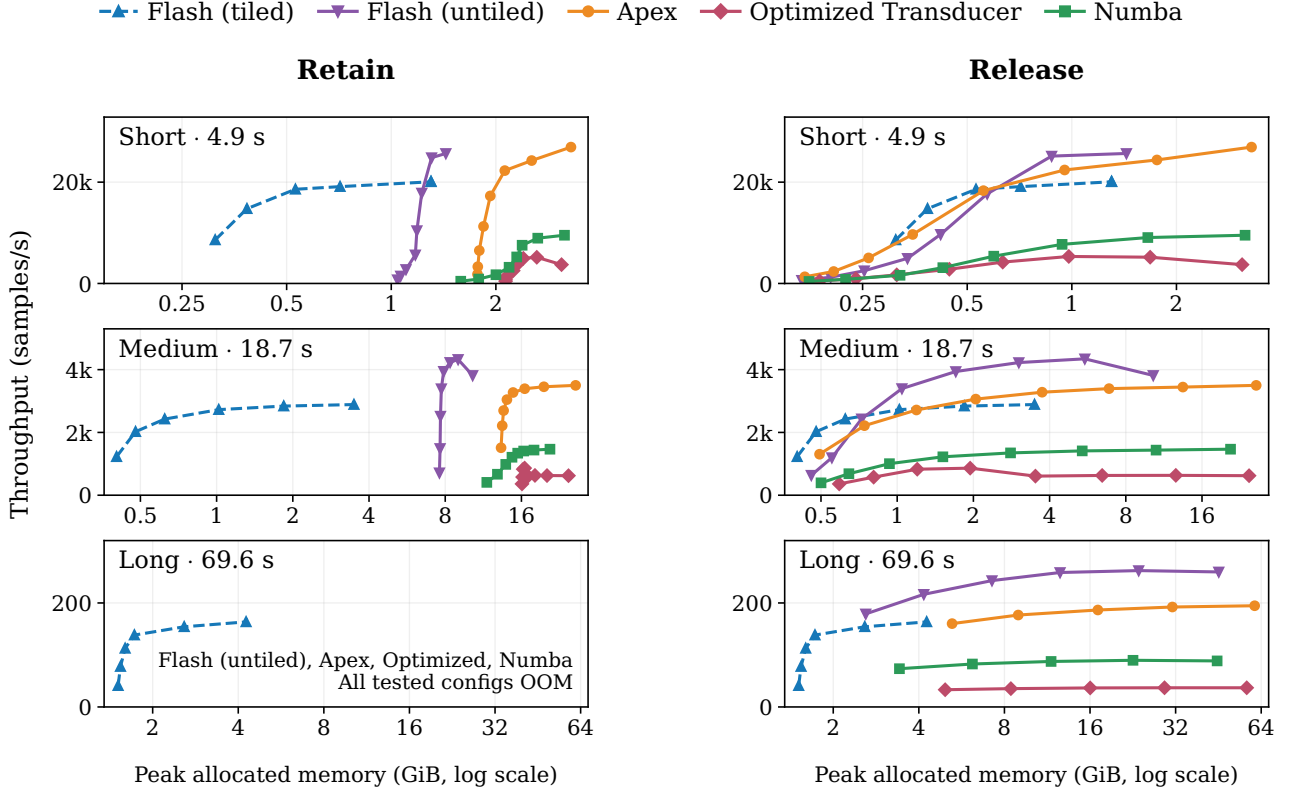}
\caption{Throughput--memory tradeoffs under retain and release scheduling. Points
correspond to tile limits for tiled Flash and sub-batch sizes for the
other implementations.}
\label{fig:3}
\end{figure*}

On H100, tiled and untiled Flash together form the entire measured
throughput--memory frontier on medium and long under both schedules (\hyperref[fig:3]{Figure~\ref*{fig:3}}).
Apex has the highest peak throughput among external implementations on all
three workloads.

Short presents a tradeoff. Using the 5\% selection rule, untiled Flash uses 60\%
less memory than Apex under retain with 7.8\% lower throughput, and 73\% less
memory under release with 6.6\% lower throughput.

On medium, the selected tiled Flash configuration reaches 2,839 samples/s at
1.84 GiB, reducing memory ninefold relative to Apex retain at 16.3\% lower
throughput. Apex release narrows this memory advantage to a factor of 3.7,
with a similar throughput gap. Untiled Flash instead improves both metrics
over Apex, providing 24--25\% higher throughput with 49--56\% less memory across
the two schedules.

On long, every tested retain configuration of untiled Flash and the external
implementations runs out of memory even with 80 GB, while all six tiled Flash
configurations fit. The selected tiled configuration reaches 163.6 samples/s
at 4.26 GiB, using fourfold less memory than Apex release at 12.3\% lower
throughput. Apex release requires a backward call for each sub-batch, whereas
tiled Flash preserves one backward call per batch. When release scheduling
is used, untiled Flash achieves 39\% higher throughput than Apex with 26\%
less memory.

\subsection{Component implementation ablation}\label{sec:5.2}

We replace compiled PyTorch \cite{ansel2024pytorch} joint and normalization components individually
and together, under untiled and tiled execution (Table~\ref{tab:2}). Compilation specializes
to each benchmark shape, so the comparison measures steady-state performance
rather than variable-length training overhead. All variants share packing and
the wavefront recurrence; untiled variants process the full batch without
sub-batching.

\begin{table}[tbp]
\centering
\caption{Throughput (samples/s) / peak allocated memory (GiB). Tiled variants use a
tile limit of 200k alignment positions.}
\label{tab:2}
\small
\setlength{\tabcolsep}{3pt}
\begin{tabular}{lrrr}
\hline
\shortstack{Components} & \shortstack{Short} & \shortstack{Medium} & \shortstack{Long} \\
\hline
\textbf{Untiled} &  &  &  \\
Compiled & 22743 / 2.80 & 3177 / 21.97 & OOM \\
Specialized joint & 21741 / 2.58 & 3205 / 20.44 & OOM \\
Specialized norm. & 26564 / 1.65 & 3878 / 11.80 & OOM \\
Both specialized & 25333 / 1.44 & 3852 / 10.27 & OOM \\
\textbf{Tiled} &  &  &  \\
Compiled & 17524 / 1.50 & 2448 / 2.16 & 124.2 / 3.49 \\
Specialized joint & 16954 / 1.35 & 2357 / 1.89 & 121.7 / 2.63 \\
Specialized norm. & 19879 / 0.86 & 2834 / 1.29 & 141.4 / 2.58 \\
Both specialized & 19135 / 0.71 & 2711 / 1.02 & 137.8 / 1.72 \\
\hline
\end{tabular}
\end{table}

Replacing compiled normalization improves tiled throughput by 13--16\%.
Adding the specialized joint reduces peak memory by a further 17--33\%.
Its throughput effect depends on the GPU, decreasing by 2.5--4.3\% on H100
but increasing on RTX 5090. Together, the specialized components remain
9--11\% faster than compiled components while using 51--53\% less memory.

Even when the joint, vocabulary projection and normalization are compiled
together, tiled Flash achieves 8--9\% higher throughput with
42--45\% less memory.

\subsection{Recomputation and release}\label{sec:5.3}

With the same specialized kernels, tiled checkpointing reduces peak memory
approximately twofold on short and tenfold on medium, at 24.5\% and 29.6\% lower
throughput, respectively. On long, all full-batch untiled variants run out of
memory, while all tiled variants fit, including those using compiled PyTorch
components.

On medium, switching untiled Flash from retain to release with 16-utterance
sub-batches reduces peak memory from 7.89 to 1.70 GiB while maintaining throughput.
On long, release with four-utterance sub-batches makes untiled execution feasible
at 243.0 samples/s and 7.23 GiB, while every tested retain configuration runs
out of memory.

The RTX 5090 experiments show the same medium- and long-workload frontier
advantage, with nearly identical peak memory for matched configurations.
Relative throughput differs across GPUs, however, as illustrated by the
specialized joint's opposite effects in \S~\ref{sec:5.2}. Complete RTX 5090 results are
available in the benchmark repository.

\section{Discussion and conclusion}\label{sec:6}

Flash RNN-T improves the measured throughput--memory tradeoff of joint-and-loss
computation while preserving the full alignment space and vocabulary. On medium
and long, tiled and untiled Flash together form the entire measured frontier on
both GPUs. On short, the selected untiled Flash configurations substantially
reduce memory relative to Apex at modestly lower throughput on H100.

The ablation explains how specialization and checkpointing contribute to this
tradeoff. Specialized normalization improves throughput over compiled PyTorch,
while the specialized joint further reduces memory with a GPU-dependent
throughput effect. Checkpointing trades additional computation for lower memory,
making the long workload feasible even with compiled components.

Using untiled Flash does not require retaining intermediates for the entire
batch. Sub-batch release makes it feasible on the long workload and faster
than tiled Flash, although it requires more memory and per-sub-batch backward
calls. Tiling instead limits joint intermediate storage to one tile and
preserves a single loss backward call, making memory reduction available without
changing the training loop.

Evaluation covers three corpus-derived batches on two GPU architectures
with one model configuration. Measurements cover joint-and-loss computation,
not end-to-end training throughput. Lattice storage still grows with sequence
length, although large joint intermediates are bounded by the tile limit.
Within this scope, Flash RNN-T offers lower-memory tiled execution and
higher-throughput untiled execution, both evaluating the full alignment
space and vocabulary.

\label{end:main}

\bibliographystyle{IEEEtran}
\bibliography{references}
\end{document}